\documentclass[lettersize,journal]{IEEEtran}
\usepackage{amsmath,amsfonts}
\usepackage{algorithmic}
\usepackage{algorithm}
\usepackage{array}
\usepackage[caption=false,font=normalsize,labelfont=sf,textfont=sf]{subfig}
\usepackage{textcomp}
\usepackage{stfloats}
\usepackage{url}
\usepackage{verbatim}
\usepackage{graphicx}
\usepackage{cite}
\usepackage[pagebackref=false, breaklinks=true, colorlinks, bookmarks=false]{hyperref}
\usepackage{booktabs}

\usepackage{tcolorbox}
\usepackage{xcolor,colortbl}
\usepackage{pifont}
\usepackage{bm}
\usepackage{subcaption}

\usepackage{color}
\usepackage{wrapfig}
\usepackage{multirow}

\begin{document}

\title{Extending Large Vision-Language Model for Diverse Interactive Tasks in Autonomous Driving}

\author{Zongchuang Zhao, Haoyu Fu, Dingkang Liang, Xin Zhou, \\   Dingyuan Zhang, Hongwei Xie, Bing Wang, and Xiang Bai
\thanks{Z. Zhao, H. Fu, D. Liang, X. Zhou, D. Zhang, and X. Bai are with Huazhong University of Science and Technology, Wuhan 430074, China. (zcuangzhao, hyfu, dkliang, xbai)@hust.edu.cn}
\thanks{H. Xie and B. Wang are with Xiaomi EV.}
\thanks{Z. Zhao and H. Fu contribute equally.}
\thanks{Corresponding author: Xiang Bai (xbai@hust.edu.cn).}
}

\makeatletter
\g@addto@macro\@maketitle

\makeatother

\maketitle

\begin{abstract}

The Large Visual-Language Models (LVLMs) have significantly advanced image understanding. Their comprehension and reasoning capabilities enable promising applications in autonomous driving scenarios. However, existing research typically focuses on front-view perspectives and partial objects within scenes, struggling to achieve comprehensive scene understanding. Meanwhile, existing LVLMs suffer from the lack of mapping relationship between 2D and 3D and insufficient integration of 3D object localization and instruction understanding. To tackle these limitations, we first introduce \textbf{NuInteract}, a large-scale dataset with over 1.5M multi-view image language pairs spanning dense scene captions and diverse interactive tasks. Furthermore, we propose \textbf{DriveMonkey}, a simple yet effective framework that seamlessly integrates LVLMs with a spatial processor using a series of learnable queries. The spatial processor, designed as a plug-and-play component, can be initialized with pre-trained 3D detectors to improve 3D perception. Our experiments show that DriveMonkey outperforms general LVLMs, especially achieving a 9.86\% notable improvement on the 3D visual grounding task. The dataset and code will be released at \url{https://github.com/zc-zhao/DriveMonkey}.

\end{abstract}


\begin{figure*} 
  \centering
\includegraphics[width=\textwidth]{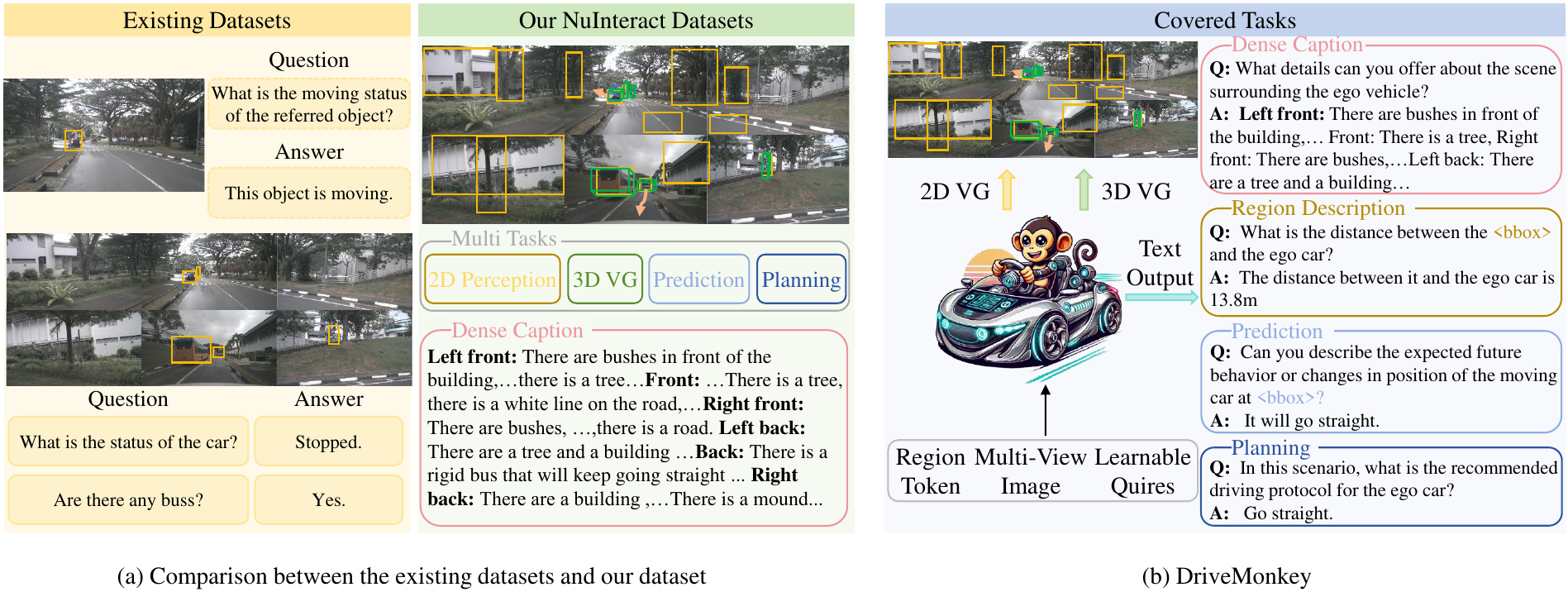}   \caption{(a) Comparison of existing datasets with NuInteract. While existing datasets use single-view images or focus on partial objects, NuInteract provides dense captions for the full scene and supports multi-object and 3D box information. (b) DriveMonkey framework utilizes multi-view images and user prompts to perform various interactive autonomous driving tasks.}
  \label{fig: overall}
  \vspace{-10pt}
\end{figure*}

\section{Introduction}

\IEEEPARstart{L}{arge} Vision-Language Models (LVLMs)~\cite{liu2023visual,chen2024internvl,achiam2023gpt,chen2024expanding_internvl25, wang2024qwen2vl,bai2025qwenvl25, team2023gemini, lin2025navcot} have made significant progress with the rapid development of Large Language Models (LLMs)~\cite{cai2024internlm2,touvron2023llama, touvron2023llama2}, presenting remarkable visual understanding and complex reasoning abilities. Recently, some efforts~\cite{ding2024holistic,nie2024reason2drive,zhou2024embodied,wang2024omnidrive, tian2024drivevlm, jiang2024senna, zhang2024wisead, huang2024drivemm} have attempted to apply LVLMs to understand autonomous driving (AD) scenarios, analyze key objects and predict the motion of ego car and other agents. They first design and curate a series of language-scene pairs for related multiple self-driving tasks (e.g., perception, critical object detection, prediction, and planning). Then, these methods transfer the excellent comprehension ability of general LVLMs to the specific driving scenes via instruction following fine-tuning, improving the interpretability and generalization for complex traffic conditions of the AD system.

Despite the rapid advances, two major issues still remain in existing works: \textbf{1) Limited environmental awareness}. As shown in Fig.~\ref{fig: overall}(a), some recent language-based autonomous driving datasets only support the single-view image~\cite{zhou2024embodied,nie2024reason2drive,marcu2024lingoqa,li2024automated_codalm}, limiting surrounding awareness and increasing safety risks. Others merely feature simplified Visual Question Answering (VQA) pairs~\cite{qian2024nuscenes, sima2024drivelm, wang2024omnidrive, wu2023language} based on multi-view images, focusing on partial objects in the scenario. For instance, while the NuScenesQA~\cite{qian2024nuscenes} and NuPrompt~\cite{wu2023language} datasets necessitate surrounding view images, their answers only either include one or two words or are related to specific objects. However, safe driving necessitates a comprehensive understanding of the ego vehicle's surrounding environment;\textbf{ 2) Insufficient 3D perception.} 3D perception is essential for reliable autonomous driving. Most current LVLMs~\cite{nie2024reason2drive,ding2024holistic} customized to the self-driving field typically lack 3D-aware capabilities due to inadequate integration of 3D geometric reasoning with natural language comprehension. Although some works~\cite{zhou2024embodied,cho2024language, wang2024omnidrive, hwang2024emma} attempt to convert 3D bounding box coordinates into textual representation for 3D perception, they face challenges in achieving satisfactory perception results due to the lack of extra 3D spatial prior (\textit{e.g.} depth and yaw) and imprecise representation of numbers.

Considering the above issues, in this paper, we first introduce a new and practical large-scale dataset, NuInteract. It comprises 239K images across 34K frames from 850 scenes, along with over 1.5M image-language pairs covering dense scene captions and diverse interactive driving tasks. As shown in Tab.~\ref{tab:data_comparison}, NuInteract offers critical advantages over existing datasets: 1) it includes dense captions detailing the environment around the ego vehicle, offering comprehensive scene information for the safe AD system; 2) it supports the 3D visual grounding (VG) task by providing 3D bounding boxes and corresponding referring expressions to localize target objects. While some datasets~\cite{deruyttere2019talk2car,sachdeva2024rank2tell} feature similar content, NuInteract not only incorporates more comprehensive information, such as multi-view and distance but also processes multiple object bounding boxes simultaneously. Moreover, as illustrated in Fig.~\ref{fig: overall}(a), our NuInteract includes a variety of tasks aligned with human logical processes (perception, prediction, and planning) during driving. Unlike previous works~\cite{sima2024drivelm,marcu2024lingoqa} that require using closed-source models or manual labeling, we design a completely automatic annotation pipeline to construct the NuInteract without any additional cost. To our knowledge, NuInteract covers most tasks that interact with user commands and information and achieves a larger scale of image-text pairs than existing datasets, enabling richer training for LVLMs.

We then benchmark representative LVLMs~\cite{liu2024improved,chen2024internvl,wang2024qwen2vl,yao2024minicpm, team2024internvl2} on the NuInteract dataset and find they struggle with 3D VG tasks. We argue that it stems from their pixel2seq paradigm~\cite{chen2021pix2seq}, where converting 3D box coordinates into text might lack spatial context and depth cues, making it challenging for models to interpret the geometric relationships essential for 3D perception. To mitigate this dilemma, we further propose \textbf{DriveMonkey}, a simple yet effective framework that can execute multiple different interactive tasks following the natural language instructions, with multi-view images as input, as shown in Fig.~\ref{fig: overall}(b). Specifically, to boost the 3D scene perception, we introduce a spatial processor, initialized from a pre-trained 3D detector~\cite{liu2022petr, li2022bevformer}, with the LVLM using a set of 3D positional-aware learnable queries. Through the special and decoupled design for 3D perception, we bridge the proposed spatial processor and LVLM to empower flexible information transmission between them. Note that this design also makes our framework a plug-and-play module, enabling our model to be updated with the various 3D object detectors and the latest LVLMs. In this way, our model not only inherits the 3D spatial perception capabilities of the object detectors but also benefits from the world knowledge of LVLMs.

Extensive experiments on the proposed NuInteract dataset demonstrate the effectiveness of our DriveMonkey. In particular, DriveMonkey significantly exceeds the general LVLMs~\cite{chen2024internvl,liu2024improved,wang2024qwen2vl} at least by 20.43\% in Precision (Pr) metric and 9.86\% in mAP on the 3D VG task. Furthermore, pre-training on our dense scene captions of NuInteract dataset yields performance improvements of 1.7\% and 0.5\% on the DriveLM dataset~\cite{sima2024drivelm} for LLaVA1.5~\cite{liu2024improved} and InternVL1.5-2B~\cite{chen2024far} separately, emphasizing the validity of our dense captions and its criticality to downstream tasks.

Our major contributions are as follows:
\begin{itemize}
    \item We present \textbf{NuInteract}, a large-scale dataset for advancing LVLMs in autonomous driving. With 239K images, 34K frames, and over 1.5M image-language pairs across 850 scenes, NuInteract provides dense captions detailing the surrounding environment and 2D/3D annotations for tasks like 2D/3D visual grounding, enabling comprehensive perception, prediction, and planning.
    \item We introduce \textbf{DriveMonkey}, a flexible framework supporting multiple interactive tasks via user prompts. Built on an LVLM, DriveMonkey integrates a plug-and-play module, spatial processor, with 3D positional-aware queries, effectively improving 3D perception ability.
    \item Extensive experiments on NuInteract demonstrate the excellent performance of DriveMonkey. Notably, after dense caption pre-training, DriveMonkey achieves significant performance improvements on various interactive tasks of NuInteract, LLaVA1.5 and InternVL1.5 on the DriveLM dataset as well, confirming the effectiveness of our dataset and training strategy.
\end{itemize}

\begin{table*}
\centering
\caption{The comparisons between our NuInteract dataset and existing language-based driving datasets. Our dataset covers most tasks that interact with user commands and information and achieves a larger scale of image-text pairs than existing datasets. 2D Perception includes 2D Region Description and 2D Visual Grounding. 3D VG denotes 3D Visual Grounding tasks. Distance information represents the absolute distance between an object and the ego vehicle.}
\footnotesize
\setlength\tabcolsep{2.1 pt} 
\definecolor{deepgreen}{rgb}{0.0, 0.5, 0.0}
\newcommand{\cmark}{{\textcolor{deepgreen}{\ding{51}}}}
\newcommand{\xmark}{\textcolor{red}{{\ding{55}}}}
\begin{tabular}{l c ccccc cccc c}
\toprule

 \multirow{2.3}{*}{Dataset} & \multirow{2.3}{*}{Reference} & \multicolumn{5}{c}{Tasks} & \multicolumn{4}{c}{Information} & \multirow{2.3}{*}{\shortstack{Image-Text\\Pairs Scale}} \\
\cmidrule(lr){3-7} \cmidrule(lr){8-11} & & Dense Caption & Prediction & 2D Perception & 3D VG &  Planning   & Multi-view & Distance & Multi-object & 3D Box \\
\midrule
    HAD~\cite{kim2019grounding} & CVPR 19 & \xmark  & \xmark & \cmark   &\xmark & \cmark  & \xmark & \xmark & \xmark & \xmark& - \\
    Talk2Car~\cite{deruyttere2019talk2car}& EMNLP 19 & \xmark  & \xmark & \cmark  & \cmark & \cmark  &  \xmark & \xmark & \xmark & \cmark & 11K\\
    DRAMA\small{-ROLISP}~\cite{ding2023hilm} &arXiv 23 & \xmark & \cmark  & \cmark &\xmark & \cmark  &  \xmark &  \xmark & \xmark & \xmark & 35K \\
    Rank2Tell~\cite{sachdeva2024rank2tell} & arXiv 23 & \xmark & \cmark & \cmark   &\xmark & \cmark  & \xmark & \xmark & \cmark & \cmark & - \\
    DRAMA~\cite{malla2023drama}& WACV 23 & \xmark  & \cmark  & \cmark &\xmark & \cmark  &  \xmark &  \xmark  & \xmark & \xmark & 100K \\
    CODA-LM~\cite{li2024automated_codalm} & arXiv 24 &\xmark & \xmark & \cmark & \xmark & \cmark & \xmark & \xmark &\xmark &\xmark & 63K \\
    DriveGPT4~\cite{xu2024drivegpt4} &RA-L 24 & \xmark & \cmark & \cmark  &\xmark & \cmark  &  \xmark &  \xmark & \cmark & \xmark & 28K \\
    Talk2BEV~\cite{choudhary2024talk2bev} & ICRA 24 &\xmark & \cmark & \cmark   &\xmark & \cmark   &  \xmark &  \cmark & \cmark & \xmark & 20K  \\
    Nuscenes-QA~\cite{qian2024nuscenes} & AAAI 24&  \xmark & \cmark  & \cmark  &\xmark & \cmark  &  \cmark & \cmark & \cmark& \xmark & 459K \\
    NuInstruct~\cite{ding2024holistic} &CVPR 24 &  \xmark & \cmark & \cmark  &\xmark & \cmark  &  \cmark & \cmark & \cmark & \xmark & 91K\\
    LingoQA~\cite{marcu2024lingoqa}  &ECCV 24  & \xmark & \xmark & \cmark   &\xmark & \cmark   & \xmark & \xmark & \xmark &  \xmark & 419K \\
    DriveLM~\cite{sima2024drivelm} &ECCV 24     & \xmark & \cmark & \cmark   &\xmark & \cmark  & \cmark & \xmark & \cmark &  \xmark & 378K \\
    Reason2Drive~\cite{nie2024reason2drive} & ECCV 24  & \xmark & \cmark  & \cmark  &\xmark & \cmark  & \xmark & \xmark & \cmark & \xmark & 600K \\
    NuPrompt~\cite{wu2023language} & AAAI 25 & \xmark & \cmark &\xmark &\cmark &\xmark &\cmark &\xmark &\cmark &\cmark &35K \\
    OmniDrive~\cite{wang2024omnidrive} & CVPR 25 & \xmark & \cmark &\cmark &\xmark &\cmark &\cmark &\xmark &\cmark &\cmark & 200K\\
    \midrule
    NuInteract \textbf{(ours)}& -  & \cmark & \cmark& \cmark & \cmark &\cmark &  \cmark &  \cmark &\cmark &\cmark &  \textbf{1.5M}\\
\bottomrule
\end{tabular}
\label{tab:data_comparison}
\vspace{-8pt}
\end{table*}

\section{Related works}

\subsection{Large Vision-Language Models}

Recent progress in LVLMs~\cite{liu2023visual,achiam2023gpt,chen2024internvl,wang2024qwen2vl} has extended the language understanding capabilities of LLMs~\cite{touvron2023llama,vicuna2023} to the visual domain. Representative models such as BLIP-2~\cite{li2023blip}, LLaVA~\cite{liu2023visual,liu2024improved, liu2024llava-next} adopt a two-stage pipeline: image-text alignment followed by instruction tuning. To enhance visual comprehension, recent works incorporate features from diverse visual encoders~\cite{radford2021clip,zhai2023sigmoid,oquab2024dinov2, zhang2024vision}, and adopt dynamic resolution strategies~\cite{li2024monkey,yao2024minicpm,liu2024llava-next,chen2024internvl} that split images into sub-regions based on their aspect ratios. Unified models like Uni-MoE~\cite{li2025uni} and Qwen2VL series~\cite{wang2024qwen2vl, bai2025qwenvl25} further extend LVLMs to multi-modal inputs (e.g., video, audio) via Mixture of Experts (MoE)~\cite{shazeer2017outrageously} or multimodal rotary position embedding (M-RoPE). Region-level understanding has been explored using token-based localization~\cite{you2023ferret,zhang2024ferret,xie2025relationlmm}, RoI-aware tokenization~\cite{ma2025groma}, and integration with segmentation networks~\cite{kirillov2023segment,cheng2022masked} through auxiliary tokens~\cite{lai2024lisa,zhang2024psalm}. VisionLLMv2~\cite{wu2024visionllmv2} introduce a super link to connect task-special decoders with LLM, completing more different tasks in a unified manner, such as image detection, segmentation, and generation, \textit{etc}.

In this paper, we aim to equip LVLMs with fine-grained 3D perception and dense environmental understanding tailored to autonomous driving scenarios.

\subsection{Language-based Driving Datasets and Models}

Several studies~\cite{qian2024nuscenes, sima2024drivelm, zhou2024embodied, li2024automated_codalm, ji2024jm3d} introduce language-based autonomous driving datasets and leverage the multi-modal capabilities of LVLMs for self-driving tasks. From the dataset aspect, some works~\cite{qian2024nuscenes,ding2024holistic, sima2024drivelm} generate VQA pairs using predefined templates~\cite{nie2024reason2drive}, SQL~\cite{ding2024holistic}, or manual annotation~\cite{marcu2024lingoqa}. DriveLM~\cite{sima2024drivelm} models graph-structured reasoning for perception, prediction, and planning. Others~\cite{deruyttere2019talk2car,malla2023drama,jin2024tod3cap} focus on object-caption pairs to aid vehicle navigation. CODA-LM~\cite{li2024automated_codalm} provides benchmarks for evaluating LVLMs in corner cases, while recent works~\cite{xie2025vlms, li2025fine} offer complex benchmarks for autonomous driving evaluation.

On the modeling side, some approaches~\cite{xu2024drivegpt4,mao2023gptdriver,wang2024omnidrive,huang2024making, hegde2025distilling, li2025generative} design end-to-end systems exploiting LVLM commonsense knowledge. ELM~\cite{zhou2024embodied} improves spatial perception and temporal modeling, while Omnidrive~\cite{wang2024omnidrive} integrates 3D spatial information for detection and prediction. Other works~\cite{zhai2024world,jiao2024lavida, huang2024drivemm} enhance VQA for autonomous driving, with MPDrive~\cite{zhang2025mpdrive} introducing marker-based prompt learning for improved perception and spatial accuracy. Some methods~\cite{wang2023drivemlm, shao2024lmdrive, fu2025orion} focus on closed-loop end-to-end driving systems. Orion~\cite{fu2025orion} uniquely bridges LVLM language space with action space for real-world interaction and motion generation.

Our NuInteract provides comprehensive scene information, enabling DriveMonkey to deliver enhanced multi-view, multi-object, and 3D-aware perception for more effective language-based driving tasks.

\subsection{Camera-based 3D object Detector}

Camera-based 3D detection~\cite{chen2024graph, wang2025physically, rahman2019notice} is a fundamental task for autonomous driving. It only requires monocular~\cite{huang2023obmo, kim2023stereoscopic} or multi-view images~\cite{li2022bevformer,liu2022petr} to perceive the environment of the ego vehicle without the sparse point cloud from LiDAR~\cite{xie2023xview, he2023sa, cai20223d}. Multi-view approaches mainly include BEV-based methods~\cite{philion2020lift, huang2021bevdet, li2022bevformer} and sparse query-based methods~\cite{liu2022petr, liu2023petrv2, chen2024graph, xiong2023cape}. BEV-based methods, such as LSS~\cite{philion2020lift} and BEVDet~\cite{huang2021bevdet}, utilize LSS operation to project multi-view features into BEV features for 3D detection, while BEVFormer~\cite{li2022bevformer} enhances it with spatial and temporal attention. Sparse query-based models like PETR series~\cite{liu2022petr, liu2023petrv2} map multi-view features to 3D space using position embeddings (PE), and CAPE~\cite{xiong2023cape} refines this by decomposing the global 3D PE into camera-view positional information. Depth supervision from the LiDAR points~\cite{li2023bevdepth,li2023bevstereo, hou2025open} and 2D auxiliary detection tasks~\cite{yang2023bevformer, jiang2024far3d} further improve 3D detection performance.

Although these 3D detectors perform well in conventional 3D detection, they cannot interact with textual information. In this paper, we organically and seamlessly combine the pre-trained multi-view camera-based 3D detectors with LVLMs to interact multi-view image features with text features, performing the 3D visual grounding task.

\begin{figure*} 
  \centering
\includegraphics[width=\textwidth]{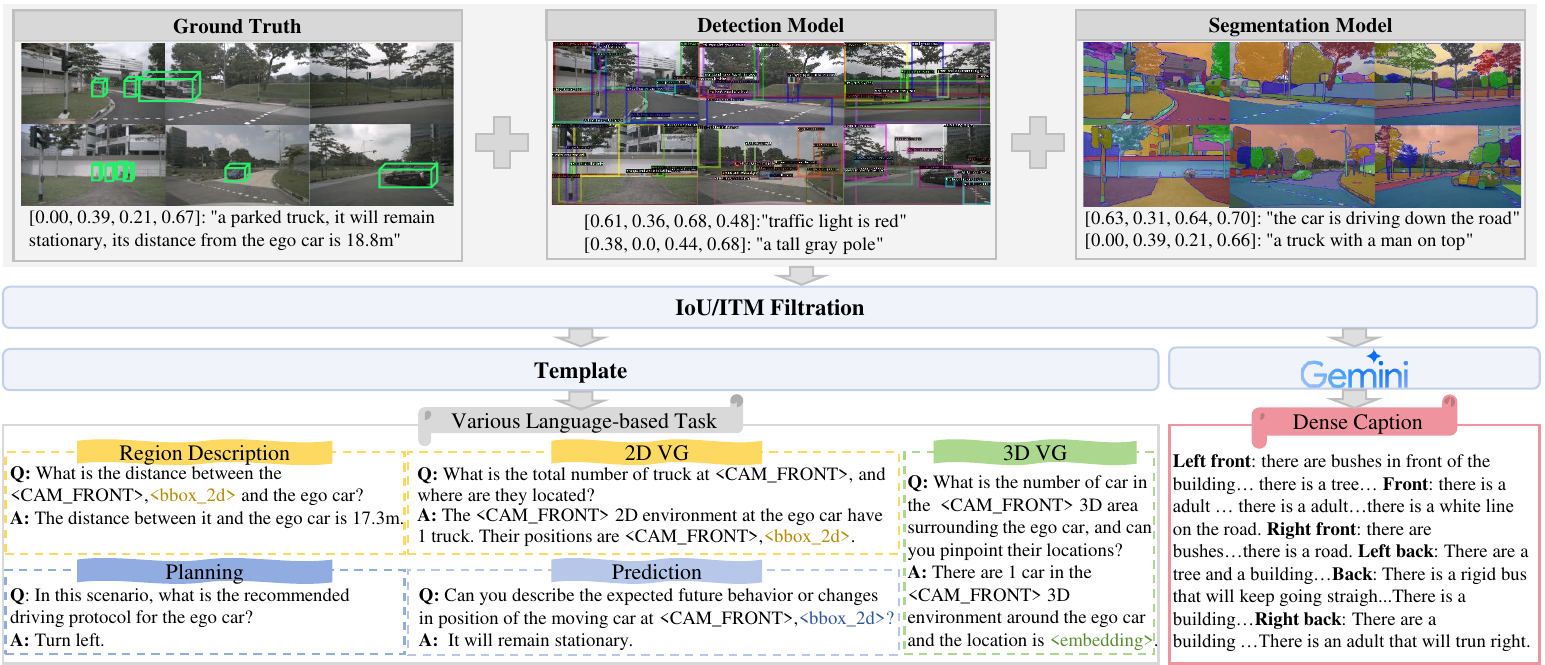}   
\caption{The pipeline of the NuInteract dataset annotation. VG denotes Visual Grounding. We collect various objects and their information from different experts, then filter them using Intersection over Union (IoU) and Image-Text Matching (ITM) criteria. The filtered objects and the corresponding information are then input into the Gemini~\cite{team2023gemini} to generate dense captions. We also use predefined templates combined with object information to create data for diverse interactive driving tasks.}
  \label{fig: anno_pipline}
  \vspace{-10pt}
\end{figure*}

\section{NuInteract Dataset}

The advancement of LVLMs significantly depends on large-scale, high-quality data~\cite{gao2023llama}, yet collecting language-related autonomous driving data remains costly and time-consuming. To fill these gaps and promote community advancement, we design a fully automated labeling pipeline (see Fig.~\ref{fig: anno_pipline}) that efficiently generates dense environmental captions and various language-based task data tailored for autonomous driving.

In Sec.~\ref{sec: Dataset_Generation}, we primarily introduce the dataset generation process, while in Sec.~\ref{sec: Dataset_Statistics}, we focus on presenting the characteristics of our dataset.

\subsection{Dataset Generation}
\label{sec: Dataset_Generation}

We conduct a generation pipeline based on the nuScenes dataset~\cite{caesar2020nuscenes}, which encompasses 850 scenes for 34K frames, to generate dense captions and various language-based interactive task data. It mainly includes collecting objects' location and the corresponding description in the current scene and the annotation process, as shown in Fig.~\ref{fig: anno_pipline}.

\textbf{Object Information Collection.} Initially, we extract Ground Truth (GT) information from nuScenes~\cite{caesar2020nuscenes}, including object distance, motion states, and attributes. Nonetheless, nuScenes annotations alone are insufficient for comprehensive scene understanding. To enrich object information, a region-to-text model~\cite{nguyen2022grit} extracts bounding boxes and corresponding object descriptions, and a segmentation model~\cite{ren2024grounded} segments and identifies objects followed by describing them using an image-caption model~\cite{li2023blip}.

However, the location and description of objects from different experts may be redundant. To mitigate redundancy and enhance object description accuracy, we first calculate the Intersection over Union (IoU) between boxes from different experts to filter duplicates\footnote{Specifically, GT boxes will be prioritized for quality, followed by the boxes from the region-to-text model due to its more accurate descriptions.}. We further employ Image-Text Matching (ITM) using BLIP2~\cite{li2023blip} to verify the consistency between image areas, objects, and corresponding descriptions, filtering out low-scoring matches. 

The entire collection process primarily relies on GT information, supplemented by expert models, ensuring the quality and authenticity of our dataset. The entire pipeline operates in a fully automated manner without additional labor or costs. 

\textbf{Annotation Process.} To further annotate scene descriptions, we prompt Gemini-Pro~\cite{team2023gemini} with a detailed template combining the information (e.g., location, category, and attributes) from all collected objects to generate dense captions, ensuring consistency and relevance to driving scenarios.

Additionally, we integrate the filtered object information into various predefined templates to create multi-view image-language pairs for different interact tasks, including 2D perception, prediction, planning, and 3D visual grounding (VG) tasks. The definitions of these tasks are as follows:
\begin{itemize}
\item \textbf{2D Perception task} includes queries about semantic and positional information of the vehicle's current scene, such as region descriptions (RD) and 2D visual grounding tasks, belonging to the region-level task. Note that we utilize object information from all expert models, not just objects in GT, to build this task data.
\item \textbf{Prediction task} involves forecasting the motion states of other vehicles, including maneuvers like turning left or right and their interactions with the ego vehicle, such as overtaking or moving away. It also attributes to the region-level task because of the same processing as the region description task. Unlike the 2D perception task, we only use GT object information to construct data.
\item \textbf{Planning task} requires the vehicle to determine appropriate actions based on available information. Here, we use the high-level command as the correct answer, including turn left, turn right, and go straight.
\item \textbf{3D Visual Grounding task} aims to localize a target object using a 3D bounding box based on a referring expression in the real driving environment. For this task, we query all 3D bounding boxes of a specific category each time in a given camera view, where single or multiple target objects are referenced by each answer.
\end{itemize}

\begin{figure*} 
  \centering
\includegraphics[width=\textwidth]{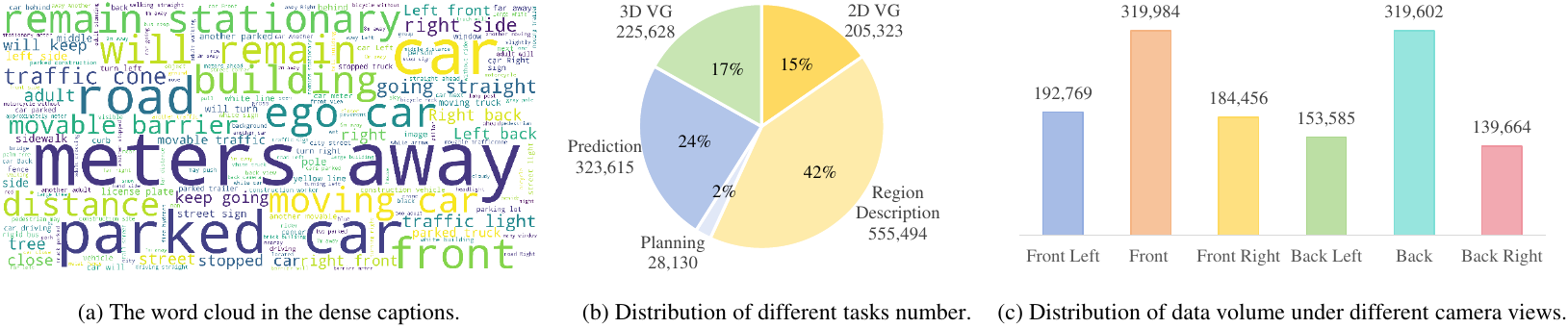}
  \caption{The statistical properties of the NuInteract dataset. VG refers to visual grounding. (a) The word frequency of dense captions. (b) The distribution of different interactive tasks. (c) The distribution of data volume under different camera views.}
  \label{fig:data_characteristics}
  \vspace{-10pt}
\end{figure*}

\subsection{Dataset Characteristics}
\label{sec: Dataset_Statistics}

We individually annotate the training and validation sets of nuScenes~\cite{caesar2020nuscenes} through the above-proposed annotation workflow, thereby constructing the training and test splits of our NuInteract. The resulting dataset encompasses 239K images (six single-view images and one surrounding view image for each frame) with high-quality dense captions and 1.3M data across diverse interactive language-based tasks excluding captions. As shown in Tab.~\ref{tab:data_comparison}, NuInteract surpasses existing driving datasets in scale, information richness, and task diversity, offering stronger support for training LVLMs.

To investigate the properties of NuInteract dataset, we first show the word frequency of the dense captions in Fig.~\ref{fig:data_characteristics}(a), leading to the following conclusions: 1) The dense captions are highly relevant to autonomous driving scenarios, focusing on multiple objects from various viewpoints, rather than solely those annotated by nuScenes~\cite{caesar2020nuscenes}, such as traffic lights, buildings, and trees. 2) High-frequency words indicate the motion states of different objects in the scene and their distances from the ego vehicle, critical for safe driving. Our dense captions fill the gap in other brief scene description datasets~\cite{wang2024omnidrive,sima2024drivelm, marcu2024lingoqa, jiang2024senna} and can serve as a valuable resource for LVLMs~\cite{wang2024qwen2vl,chen2024internvl,chen2024expanding_internvl25, liu2024llava-next} in the self-driving domain. 

We then count the distribution of different interactive tasks, as shown in Fig.~\ref{fig:data_characteristics}(b). The distribution is generally balanced across tasks, except for planning tasks, where each frame has only one planning trajectory. The 2D perception task, including region descriptions and 2D VG task, constitutes most of the dataset due to the integration of various expert models. Additionally, the prediction and 3D VG tasks are distributed as 24\% and 17\%, respectively.

Compared with other single-view datasets~\cite{marcu2024lingoqa,nie2024reason2drive}, NuInteract utilizes the multi-view information. Fig.~\ref{fig:data_characteristics}(c) illustrates the data volume distribution under different camera perspectives. We observe that the dataset primarily focuses on information from the front and back views, while other views are evenly distributed. It stems from the fact that the front and rear views typically encompass more salient targets during driving, as annotated in nuScenes~\cite{caesar2020nuscenes}.

\section{Method}

\begin{figure*} 
  \centering
  \includegraphics[width=\textwidth]{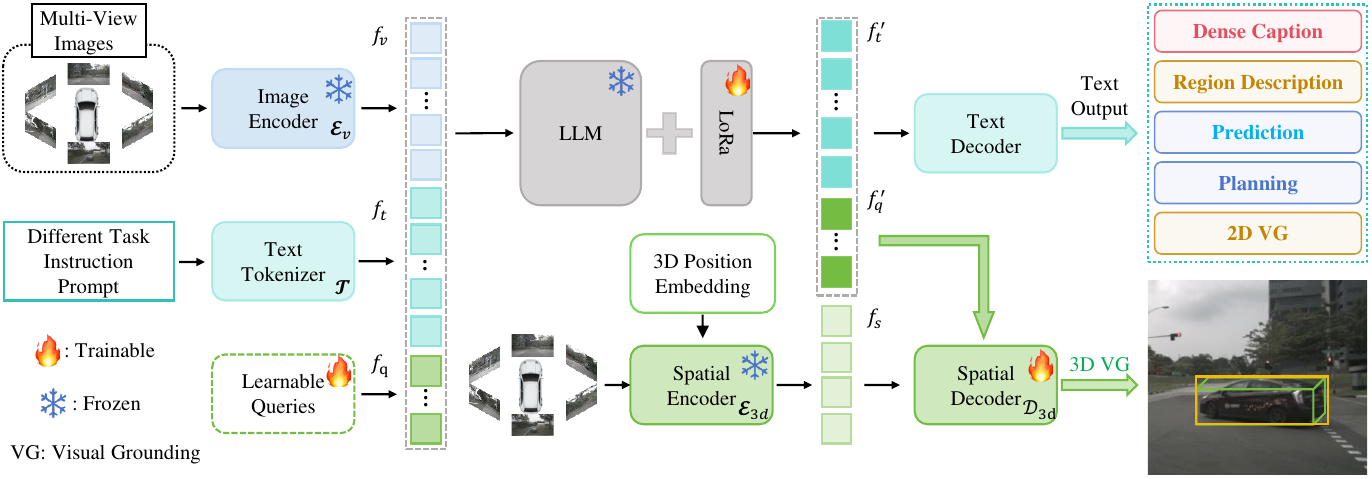}
  \caption{\textbf{The overall architecture of our DriveMonkey.} The model first encodes input instruction prompts and multi-view images into embeddings. These embeddings and a group of learnable queries are fed into a Large Language Model  (LLM). The output text tokens are used to generate associated language outputs. The spatial decoder receives the spatial feature from the spatial encoder, along with the learnable queries processed by the LLM, to detect the corresponding 3D object location.}
  \label{fig: architecture}
  \vspace{-10pt}
\end{figure*}

In this section, we provide a detailed overview of the DriveMonkey framework. Fig.~\ref{fig: architecture} illustrates the overall architecture of DriveMonkey. The input information includes multi-view images and user instruction prompts. Image encoder and text tokenizer encode the image and text information into image and text token embeddings, respectively. These features, concatenated with a series of learning queries, are subsequently processed by the LLM. Then, the output text tokens are used to generate the related answers, and the spatial decoder accepts the spatial features from the spatial encoder and learnable queries processed by the LLM, producing the final 3D object location. This decoupled design allows our model to flexibly handle diverse interactive tasks with different forms of inputs and outputs in a unified manner.

In Sec.~\ref{sec: Method_Unifying Input Format}, we detail the input format of our model. Sec.~\ref{sec: Method_DriveMonkey Framework} focuses on the architecture of DriveMonkey. Sec.~\ref{sec: Training Objectives} introduces the training objective of our method.

\subsection{Unifying Input Format}
\label{sec: Method_Unifying Input Format}
Developing a unified model to handle diverse interactive tasks of self-driving scenarios is challenging due to the distinct input formats of these tasks, such as the input image, text, and 2D/3D coordinates. To tackle this problem, we propose a unified expression approach, forming the foundation for the development of DriveMonkey.

\textbf{Scene Understanding.} For the scene understanding task, such as dense caption generation and the planning task based on the comprehensive surrounding information of the ego car, given input text embeddings ${f}_t \in \mathbb{R}^{L \times C}$, the model takes input multi-view images $\mathcal{I} =\{\mathcal{I}_{i}\}_{i=1}^{n}$, and output the text description. $L$, $C$ and $n$ are the length of the input text tokens, the channel dimension of LLM and the number of views.

Meanwhile, to enable the model to differentiate image features from various views and establish connections between them, we design a simple yet effective surround-view prompt tailored for driving scenarios: \texttt{The <image> <image> <image> <image> <image> <image> present an overview of the surrounding scene of ego vehicles, sequentially from the front left, front, front right, back left, back, and back right perspectives of the ego vehicle}, where \texttt{<image>} is a placeholder and will be replaced by the image embedding during generation.

\textbf{2D region-related Tasks.} For the 2D region-related tasks, including 2D perception and prediction, it is essential to process 2D coordinates and regional information for the model’s accurate understanding and generation of region descriptions. For the given 2D bounding box, we follow the processing method of the current LVLMs with outstanding region understanding ability, normalizing the coordinates (within the range [0, 1000)), and then transform it into a specified string format: ``$(X_{tl}, Y_{tl}), (X_{br}, Y_{br})$".

\textbf{3D Visual Grounding.} For the 3D visual grounding (VG) tasks, in addition to text token ${f}_t$ and multi-view images $\mathcal{I}$, the model further takes extra learnable queries $f_q \in \mathbb{R}^{Q \times C}$ as inputs and outputs the corresponding 3D bounding box, where $Q$ denotes the number of queries. To maintain consistency with the text input format, we introduce a placeholder, `\texttt{<embedding>}', replaced by the learnable queries when performing the 3D VG task.

\subsection{DriveMonkey Framework}
\label{sec: Method_DriveMonkey Framework}
The overall architecture of DriveMonkey is shown in Fig.~\ref{fig: architecture}. It mainly contains two parts: LLaVA-like LVLM and spatial processor with encoder and decoder.

\textbf{Pre-trained LVLMs.} We adopt pre-trained LLaVA-like models as the LVLM, consisting of an image encoder $\mathcal{E}_{v}$, a projector $\mathcal{P}$, and a LLM. Given the multi-view images $\mathcal{I}$, these images are processed through the image encoder and then mapped to language space using the projector, resulting in multi-view features ${f}_v \in \mathbb{R}^{(n\times M) \times C}$. This is formulated as:
\begin{equation}
        {f}_v = \mathcal{P}([\mathcal{E}_{v}(\mathcal{I}_1);\cdots; \mathcal{E}_{v}(\mathcal{I}_N)]),~~\mathcal{E}_{v}(\mathcal{I}_i) \in \mathbb{R}^{M \times C},
\end{equation}
where $M$ is the feature-length of a single image. The LLM accepts these features combined with input text embeddings and output text prediction $\hat{y}_t$ as:
\begin{equation}
    \hat{y}_t = \mathit{LLM}({f}_v, {f}_t ).
\end{equation}
Note that we use pre-trained LVLMs following previous methods~\cite{rasheed2024glamm, zhang2024psalm}. We comply with the same pipeline~\cite{wang2024qwen2vl, chen2024internvl} for all tasks except for the 3D visual grounding task, without further modification.

\textbf{Spatial Processor.} 
Pioneering LVLMs~\cite{wang2024qwen2vl,chen2024expanding_internvl25} often struggle to perceive 3D spatial information from multi-view images due to the lack of 3D prior input (\textit{e.g.} depth and yaw information). In contrast, camera-based 3D object detectors~\cite{liu2022petr,li2022bevformer,xiong2023cape} demonstrate strong 3D perception capabilities from multi-view images and can produce accurate 3D bounding boxes, but they cannot process textual input to perform 3D VG tasks. To overcome this limitation, our DriveMonkey integrates a 3D detector as a spatial processor to provide 3D spatial priors and combines it with an LLM. Specifically, we adopt the image backbone and position encoder of the 3D object detector~\cite{liu2022petr} as our spatial encoder $\mathcal{E}_{3d}$ and employ its object decoder as our spatial decoder $\mathcal{D}_{3d}$. Additionally, we introduce a set of learnable queries $f_q$ to serve as a communication bridge between the 3D detector and the LLM.

Unlike specialized detectors that feed learnable queries directly into the object decoder, we first update these queries within the LLM to accumulate the textual information and then map updated queries into the space of the spatial decoder via a two-layer MLP. This technique establishes an effective communication bridge and facilitates information transmission between the LVLM and the spatial processor. As illustrated in the green part of Fig.~\ref{fig: architecture}, DriveMonkey performs the 3D VG task as follows:
\begin{equation} \begin{aligned}
f_q^{'} = \mathit{LLM}(f_v ,f_t, f_q)&, \quad f_s = \mathcal{E}_{3d}(\mathcal{I},P_{3d} ), \\
\hat{c}_{3d}, \hat{b}_{3d} =\mathcal{D}_{3d}&(f_s, MLP (f_q^{'}) ),
\end{aligned} \end{equation}
where $f_q^{'}$ is the group of learnable queries updated by LLM. $P_{3d}$, $f_s$, $\hat{c}_{3d}$, and $\hat{b}_{3d}$ represent the 3D Position Embedding, spatial feature, predicted categories, and coordinates of the 3D bounding box, respectively.

\textbf{The advantage of decoupled design.} By leveraging the learnable queries, we seamlessly integrate the LVLM with pre-trained 3D detectors, endowing the LVLM with 3D perception capabilities. This not only enhances LVLM to understand and reason about 3D spatial information but also enables camera-based 3D detectors to interact with user instructions. Meanwhile, this design allows seamless integration with various camera-based 3D object detectors~\cite{liu2022petr,li2022bevformer,xiong2023cape} and LVLMs utilizing different image encoders~\cite{radford2021clip,zhai2023sigmoid} and LLMs, demonstrating the flexibility of our framework.

\subsection{Training Objectives}
\label{sec: Training Objectives}
In this subsection, we detail the training stage and the specialized losses designed for DriveMonkey to handle multiple tasks effectively.

Based on the proposed unified input format and simple yet effective framework, we employ a two-stage training strategy for DriveMonkey, following the general LVLMs~\cite{liu2023visual,team2024internvl2}. In the first stage, we merely train the projector $\mathcal{P}$ using the dense captions provided by NuInteract while freezing the parameters of the image encoder and LLM in DriveMonkey. This enables multi-view image features to be mapped into language space, equipping our model to understand the driving scenarios comprehensively. Note that the spatial processor is not introduced yet in this stage. For the second stage, we integrate the spatial processor into the model and perform mixed multi-task fine-tuning, where we fine-tune DriveMonkey based on the NuInteract all-train dataset, including the dense caption and other interactive tasks. During this stage, the spatial decoders undergo training, and the LLM is fine-tuned using LoRA~\cite{hu2021lora}.

During the first training stage, we only adopt the next token prediction loss $\mathcal{L}_{text}$ within the LLM. At the second period, we combine the 3D detection loss and the text prediction loss, as $\mathcal{L} = \mathcal{L}_{3d} + \mathcal{L}_{text}$. Specifically, for the 3D VG task, we follow the specific camera-based 3D detector and adopt focal loss~\cite{ross2017focal} for classification alongside L1 loss for 3D bounding box regression. The specialized loss is summarized as:
\begin{equation}
    \mathcal{L}_{3d} = \lambda \mathcal{L}_{cls}(c_{3d}, \sigma(\hat{c}_{3d})) + \gamma \mathcal{L}_{reg}(b_{3d}, \sigma(\hat{b}_{3d})),
\end{equation}
where $c_{3d}/\hat{c}_{3d}$ and $b_{3d}/\hat{b}_{3d}$ represent the ground-truth and predicted categories and 3D bounding box coordinates, respectively. The function $\sigma$ refers to the Hungarian algorithm~\cite{kuhn1955hungarian} for label assignment between ground-truths and predictions. We set $\lambda = 2$ and $\gamma = 0.25$ to balance the different loss components. For other tasks, we follow the first training stage and only utilize the next prediction token loss $\mathcal{L}_{text}$.

\section{Experiments}
\subsection{Implementation Details}

All experiments are conducted on eight NVIDIA A800 GPUs with 80GB of memory. InternVL2-8B~\cite{chen2024internvl} serves as the primary baseline. Input images are resized to $448 \times 448$ pixels to ensure consistent dimensions before the image encoder processing. Unless otherwise specified, the spatial processor is initialized by pre-trained PETR and connects with LVLM through 30 learnable queries, namely $Q = 30$.

In the first stage, we train DriveMonkey for 1 epoch, with a total batch size of 256. We use AdamW~\cite{loshchilov2017decoupled} as the optimizer, applying a weight decay of 0.1, and a cosine annealing scheduler with the max learning rate of $2\times10^{-5}$. For the second stage, the model undergoes 1 train epoch with a peak learning rate of $4\times10^{-5}$ and a total batch size of 128, using the same optimizer. We set a LoRA rank of 128 and an alpha of 256 for LLM fine-tuning.

\noindent\textbf{Dataset.} We primarily evaluate the DriveMonkey on the constructed NuInteract test set, obtained from the nuScenes~\cite{caesar2020nuscenes} validation set based on our proposed generation workflow in Sec.~\ref{sec: Dataset_Generation}. With the introduction of expert models in the workflow, partial objects are excluded from the GT information in the nuScenes dataset. Note that partial scenes from our dataset test set are selected to ensure balanced planning commands that reflect planning capabilities in the planning tasks.

In addition to NuInteract, we also utilize the DriveLM-nuScenes dataset~\cite{sima2024drivelm} to assess the validity of our dense caption. It consists of 4,871 driving frames with multi-view images, and question-answering data. Each frame contains 91.4 question-answer pairs on average, covering multiple driving tasks such as perception, prediction, and planning. Unless otherwise specified, when we use DriveLM, we train the models on its training set and report metrics based on its test set.

\noindent\textbf{Metric.} For region description and prediction tasks (the definition of these tasks lists in Sec.~\ref{sec: Dataset_Generation}), we use a suite of metrics, including CIDEr~\cite{vedantam2015cider}, BLEU~\cite{papineni2002bleu}, and ROUGE-L~\cite{lin2004rouge}, with $\log_{10}{\left (\text{CIDEr}+1\right )} $ rescaling to normalize CIDEr scores between 0 and 1. In the 2D visual grounding (VG) task, mean IoU (mIoU) is calculated between predictions and ground truth with the threshold of 0.5. We also use the mAP and F1 score following the COCO protocol~\cite{lin2014coco}. In the planning task, accuracy is used as the metric. 

For the 3D VG task, grounding performance is evaluated using Precision (Pr), mAP, and F1 score. Since general LVLMs lack confidence scores, we follow ELM~\cite{zhou2024embodied} and apply Hungarian matching~\cite{kuhn1955hungarian} for one-to-one comparisons of predicted and ground truth 3D boxes, as defined below:
\begin{equation}
\text{Pr}@k = \frac{1}{N_{gt}} \sum_{i=1}^{N_{gt}} \left (\left\| \hat{b}_{3d} - b_{3d} \right\|_2 ^{i} < k \right ), k=\left \{ 0.5,1,2,4 \right \},
\label{eq:pr3d}
\end{equation}
where $\left\| \cdot  \right\|_2$ is the L2 distance, and $N_{gt}$ is the number of ground truth (GT) bounding boxes. For mAP, we follow the nuScenes~\cite{caesar2020nuscenes} protocol, treating boxes as true positives when the center distance between predicted and GT boxes is less than 0.5. As for the F1 score, we first calculate per-pair IoUs between ground truth and predicted bounding boxes and then apply Hungarian matching~\cite{kuhn1955hungarian} for one-to-one comparisons of predicted and ground truth 3D boxes. After obtaining the optimal match, we take pairs with IoUs higher than the threshold as True Positives (TP), and then we can obtain precision and recall, calculating the F1 score.

\begin{table*}
    \centering
    \footnotesize
    \setlength\tabcolsep{1.8mm} 
    \renewcommand{\arraystretch}{1.3} 
    \caption{Performance of various models on the NuInteract test set. All LVLMs are fine-tuned on NuInteract train data. \textbf{Bold} emphasizes top method; \underline{underline} marks the runner-up. $\star$ indicates not using dynamic resolution and only utilizing the global image. ${\dagger}$ implies directly using specialized 3D detectors for detection. RD: Region Description; Pre: Prediction; VG: Visual Grounding; Plan: Planning; B: BLEU; R\_L: ROUGR\_L; C: CIDER; Pr: Precision; Acc: Accuracy.}
    \begin{tabular}{l c c   c c c   c c c   c c c   c  c c}
    \toprule
    
    \multicolumn{1}{l}{\multirow{2.2}{*}{Model}} & \multicolumn{1}{c}{\multirow{2.2}{*}{Years}} &  \multicolumn{1}{c}{\multirow{2.2}{*}{LLM}} & \multicolumn{3}{c}{2D RD \& Pre $\uparrow$} & \multicolumn{3}{c}{2D VG $\uparrow$} & \multicolumn{3}{c}{3D VG $\uparrow$} & Plan $\uparrow$ & \multicolumn{1}{c}{\multirow{2.2}{*}{Avg.}} \\
    \cmidrule(lr){4-6} \cmidrule(lr){7-9} \cmidrule(lr){10-12} \cmidrule(lr){13-13} & & & B & R\_L & C &mAP &F1 & MIoU  & Pr &mAP &F1 & Acc & \\   
    \midrule
    LLaVA1.5~\cite{liu2024improved} &2024 &Vicuna-7B  &64.23 &76.79	&74.82	&0.10	&0.16 &14.31  &6.51 &5.33 & 3.12	&36.20 & 28.16\\
    MiniCPM-V 2$^{\star}$~\cite{yao2024minicpm} &2024 & MiniCPM-2B & 47.43 & 63.16 &69.88 &0.11 &0.13 & 13.34 &0.97	&1.55 &0.86 & 36.69 & 23.41 \\
    MiniCPM-V 2.6$^{\star}$~\cite{yao2024minicpm} & 2024 & Qwen2-7B & 47.92 &69.11 &70.20 &0.36 &0.49 & 18.74  &1.97 & 1.61  &0.93 & 36.42 & 24.78  \\
    InternVL1.5-2B$^{\star}$~\cite{chen2024far} &2024 &InternLM2-2B  &67.14 &81.10 &\underline{79.83}	&14.74	&17.64 &55.43 &28.05 &21.73  &12.92 & \underline{53.96} & 43.25 \\
    InternVL1.5-4B$^{\star}$~\cite{chen2024far} &2024 &Phi3-4B  &66.63	&80.64	&79.24	&14.27	&17.60 &53.52 &25.14	&19.46 &11.63 & 40.25 & 40.84\\
    Qwen2VL~\cite{wang2024qwen2vl} &2024 & Qwen2-2B & \textbf{67.92} & 80.24 & 78.51 & 17.11 & 20.87 & 57.24 &12.82  &10.20 &6.12 & 45.59 &39.66\\
    Qwen2VL~\cite{wang2024qwen2vl} &2024 & Qwen2-7B & 66.65 & 78.57 & 77.97 & 16.06 & 20.04 & 55.51 &20.64  &16.26 &9.82 & 49.33 & 41.09\\
    InternVL2-1B$^{\star}$~\cite{team2024internvl2} &2024 &Qwen2-0.5B  &66.89	&81.00	&79.59	&16.70	&20.21 &55.94 &23.36	&18.35 &10.94 & 44.08 & 41.71\\
    InternVL2-2B$^{\star}$~\cite{team2024internvl2} &2024 &InternLM2-2B  &66.77	&80.87	&79.62	&16.12	&19.49 &55.29 &27.83	&21.09 &12.58 & 44.61 & 42.43\\
    InternVL2-4B$^{\star}$~\cite{team2024internvl2} &2024 &Phi3-4B  &66.88	&80.76	&79.29	&19.14	&23.47 &59.07 &25.28	&20.12 &11.97 & 40.43 & 42.64\\
    InternVL2-8B$^{\star}$~\cite{team2024internvl2} &2024 &InternLM2.5-7B  &67.32	&\textbf{81.39}	&\textbf{80.01}	&\textbf{20.61}	&\textbf{25.24} &\textbf{61.90} &31.47	&24.67 &14.70 & 46.93 & \underline{45.42} \\
    \midrule

    Bevformer$^{\dagger}$~\cite{li2022bevformer} &2022  &- &\multicolumn{6}{c}{\multirow{3}{*}{Unsupported}}	&44.50 &23.69 &15.76 & \multicolumn{2}{c}{\multirow{3}{*}{Unsupported}} \\
    PETR$^{\dagger}$~\cite{liu2022petr} &2022 &- & & & & & &	&\textbf{55.80} &31.34 &20.58 & &\\
    CAPE$^{\dagger}$~\cite{xiong2023cape} &2023 &- & & & & & &	&\underline{55.02} &\underline{32.94} &\textbf{21.33} & &\\
    
    \midrule
    \rowcolor[RGB]{230, 242, 255} DriveMonkey \textbf{(ours) } & 2025 &InternLM2.5-7B &\underline{67.50}	&\underline{81.15}	&79.79	&\underline{19.47}	&\underline{24.02} &\underline{59.36} &51.90	&\textbf{34.53} &\underline{20.86} & \textbf{82.64} & \textbf{52.12}\\
    \bottomrule
    \end{tabular}
    \label{tab:sota_table}
\end{table*}
 
\begin{table*}
    \centering
    \footnotesize
    \caption{Analysis of Dense Caption Pretraining. ``Caption Pretrain" denotes using our caption to pre-train. RD: Region Description; Pre: Prediction; VG: Visual Grounding; Plan: Planning. Metrics: B: BLEU; R\_L: ROUGE-L; C: CIDER; FS: Final Score; Acc: Accuracy; S: SPICE; Pr: Precision; mIoU: mean Intersection over Union; mAP: mean Average Precision.}
    
    \begin{minipage}[t]{0.40\textwidth}
        \centering
        \setlength\tabcolsep{2pt} 
        \renewcommand{\arraystretch}{1.0} 
        \text{(a) DriveLM Test} \\[0.5em] 
        \begin{tabular}{l  c c c c c c c}
        \toprule
        Method &   Acc & GPT & B-1 & MAT & R\_L & C & FS \\
        \midrule
        LLaVA1.5~\cite{liu2024improved}  & 64.3 & 51.3 & 74.9 & 33.7 & 72.5 & \textbf{14.8} & 49.5 \\ 
        + Caption Pretrain  & \textbf{69.1} & \textbf{52.1} & \textbf{76.3} & \textbf{35.6} & \textbf{73.3} & 13.2 & \textbf{51.2} \\
        \midrule
        InternVL1.5-2B~\cite{chen2024far} & 65.1 & 55.5 & \textbf{73.1} & \textbf{36.1} & 71.2 & \textbf{15.8} & 51.7 \\ 
        + Caption Pretrain & \textbf{69.9} & \textbf{55.5} & 72.6 & 34.3 & \textbf{71.2} & 14.4 & \textbf{52.2} \\ 
        \bottomrule
        \end{tabular}
    \end{minipage}
    \hfill
    \begin{minipage}[t]{0.57\textwidth}
        \centering
        \setlength\tabcolsep{2.0pt} 
        \renewcommand{\arraystretch}{1.25} 
        \text{(b) NuInteract Test} \\[0.5em] 
        \begin{tabular}{l  c c c  c c c   c c c }
        \toprule
        \multicolumn{1}{l}{\multirow{2.3}{*}{Model}} &  \multicolumn{3}{c}{2D RD\& Pre} & \multicolumn{3}{c}{2D VG} & \multicolumn{3}{c}{3D VG} \\
         \cmidrule(lr){2-4} \cmidrule(lr){5-7} \cmidrule(lr){8-10}  &B  &R\_L  & C  &mAP &F1 & MIoU  & Pr &mAP &F1 \\
        \midrule
        DriveMonkey  & 67.02  &80.45 & 79.20 &16.96 &20.56 &56.46 &51.47 &33.95 & 20.26 \\ 
        + Caption Pretrain & \textbf{67.21}    & \textbf{80.45}   & \textbf{79.31}   & \textbf{17.07}   & \textbf{20.60}  & \textbf{56.92}  & \textbf{51.66} & \textbf{34.26}  & \textbf{20.48} \\
        \bottomrule
        \end{tabular}
    \end{minipage}
    \label{tab:caption_ablation}
    \vspace{-10pt}
\end{table*}

\subsection{Main results}
We conduct a comparative analysis against several advanced general LVLMs~\cite{liu2024improved,chen2024internvl,yao2024minicpm, wang2024qwen2vl, team2024internvl2} in the image domain. They universally convert 2D/3D bounding box coordinates into textual format, treating all tasks as text-generation tasks within an auto-regressive paradigm. Additionally, we benchmark our framework against some traditional 3D detectors on the 3D VG task. Since they lack inherent 3D VG capabilities, we implement a proxy evaluation wherein candidate bounding boxes are first selected based on confidence scores and projection camera views, followed by calculating corresponding performance metrics.

As reported in Tab.~\ref{tab:sota_table}, DriveMonkey significantly outperforms previous LVLMs in terms of average metrics across all tasks on our NuInteract test dataset. Our model achieves comparable performance on the 2D tasks to the internVL2-8B~\cite{team2024internvl2} sharing the identical image encoder and LLM, demonstrating the effective knowledge transfer capabilities of our framework. Additionally, DriveMonkey surpasses existing LVLMs and achieves comparable performance to the specialized 3D detectors for 3D VG, setting a new benchmark for precision and mAP. Specifically, Drivemonky outperforms existing general LVLMs by a significant margin of 9.86\% on mAP, rivaling the performance of advanced traditional 3D detectors. This phenomenon indicates that our framework synergistically integrates LVLM and pre-trained 3D detectors, seamlessly inheriting their strengths. Notably, in critical planning tasks for autonomous driving, our model also achieves superior accuracy, outperforming InternVL2-8B by a large margin. DriveMonkey demonstrates exceptional versatility in processing diverse input modalities and generating various output formats, while comprehensively addressing multiple interactive tasks essential for autonomous driving applications.

\subsection{Dataset Analysis}
We evaluate the validity of the proposed dense caption on the DriveLM dataset and NuInteract dataset, with results summarized in Tab.~\ref{tab:caption_ablation}. Compared to directly fine-tuning on DriveLM, models that undergo our dense caption pre-training demonstrate superior performance. Specifically, LLaVA1.5 and InternVL1.5-2B achieve gains of 1.7\% and 0.5\%, respectively, in the final score on the DriveLM test set. Notably, both models exhibit a 5.8\% improvement in accuracy, which can be attributed to our dense captions detailing the surrounding environment of the ego vehicle. Furthermore, the dense caption consistently boosts the performance of DriveMonkey across all tasks within the NuIteract dataset. Compared to direct fine-tuning, we observe improvements of 0.19\% in BLEU score for the 2D region description and prediction task, 0.11\% in mAP and 0.46\% in mIoU for the 2D VG task, and 0.31\% in mAP and 0.22\% in F1 score for 3D VG task. These results collectively demonstrate the effectiveness of our dense caption pre-training in facilitating a multimodal understanding of autonomous driving scenarios and further validate the high quality and utility of the NuInteract dataset.

\subsection{Ablation Study}
To better and faster demonstrate the flexibility and effectiveness of our framework, unless otherwise specified, the following experiments use InternVL2-2B as the base LVLM.

\begin{table}
    \centering
    \footnotesize
    \setlength\tabcolsep{2.7mm}
    \caption{The 3D Visual Grounding performance of DriveMonkey with diverse LVLMs.}
	\begin{tabular}{l c   c c c }
	\toprule
    LVLMs & LLM & Pr &mAP &F1 \\
	\midrule
    MiniCPMV-2~\cite{yao2024minicpm} &MiniCPM-2B  & 31.80 & 11.39 &6.45 \\
    MiniCPMV-2.6~\cite{yao2024minicpm} &Qwen2-7B  &29.88 &9.71 &5.51 \\
    InternVL1.5-2B~\cite{chen2024far} &InternLM2-2B   &50.19 &32.88 & 19.72 \\
    InternVL1.5-4B~\cite{chen2024far} &Phi3-4B  &49.11 &24.07 & 14.39 \\
    
    Qwen2VL-2B~\cite{wang2024qwen2vl} &Qwen2-2B &33.31 &16.37 & 9.76  \\
    
    Qwen2VL-7B~\cite{wang2024qwen2vl} &Qwen2-7B   &23.79 &11.06 & 6.75 \\
    InternVL2-1B~\cite{team2024internvl2} &Qwen2-0.5B  &50.67	&33.29 &19.94 \\
    InternVL2-2B~\cite{team2024internvl2} &InternLM2-2B  &51.66	&34.26 &20.48 \\
    InternVL2-4B~\cite{team2024internvl2} &Phi3-4B  &50.75	&31.67 &19.24  \\
    InternVL2-8B~\cite{team2024internvl2} &InternLM2.5-7B  &\textbf{51.90}	& \textbf{34.53} & \textbf{20.86} \\
    \bottomrule
\end{tabular}
\label{tab:llm_comparison}
\end{table}

\begin{table}[!t]
\centering
    \footnotesize
    \setlength{\tabcolsep}{1.2mm}
    \caption{The effect of various 3D detectors as the spatial processor on 3D Visual Grounding tasks. ${\dagger}$ implies only using current frame information instead of default multiple frames.}
	\begin{tabular}{l c c c c c c c}
	\toprule
 3D Detector & Pr@0.5 & Pr@1 & Pr@2  & Pr@4 & Pr &mAP  &F1 \\
	\midrule
    BEVFormer$^{\dagger}$~\cite{li2022bevformer} & \textbf{25.70} &  \textbf{43.09} & 61.28 & 76.90 & \textbf{51.74} & 30.87 & 17.85 \\
    CAPE~\cite{xiong2023cape}& 10.79  & 27.29 &  51.27 & 73.24 & 40.65  & 31.51& 19.00 \\
    PETR~\cite{liu2022petr} & 19.95  & 40.78  & \textbf{64.31}  & \textbf{81.61}  & 51.66 & \textbf{34.26} &\textbf{20.48} \\
    \bottomrule

\end{tabular}
\label{tab:3d_component}
\end{table}

\begin{table}[!t]
\centering
    \footnotesize
    \caption{Ablation study of 3D Visual Grounding performance on learnable query number design.}
    \label{tab:query_num_ablation}
    \setlength{\tabcolsep}{2.0mm} 
    \begin{tabular}{c c c c c c c c} 
        \toprule
        Query & Pr@0.5 & Pr@1 & Pr@2 & Pr@4 & Pr &mAP &F1 \\
        \midrule
        10 & 16.25 & 33.75 & 53.87 & 71.91 & 43.94 & 33.10 & 19.89\\
        30 & 19.95 & 40.78 & 64.31 & 81.61 & 51.66 & \textbf{34.26} & \textbf{20.48} \\
        100 & 23.11 & 46.10 & 70.19 & 84.44 & 55.96 & 23.02 & 13.90 \\
        500 & \textbf{24.33} & \textbf{49.38}  & \textbf{74.09} & \textbf{86.09} & \textbf{58.47}  & 21.32 & 12.94 \\
        900 & 23.63 & 46.96 & 72.04 & 85.46 & 57.02 & 7.12&4.52  \\
        \bottomrule
    \end{tabular}
    \vspace{-10pt}
\end{table}

\textbf{DriveMonkey with various LVLMs.} Our modular framework allows for the integration of diverse LVLMs with the spatial processor. To explore the influence of different LVLM backbones, we evaluate DriveMonkey coupled with diverse LVLMs on the 3D VG task. As shown in Tab.~\ref{tab:llm_comparison}, we can make the following observations: 1) When incorporating MiniCPMV~\cite{yao2024minicpm} series, our framework struggles to utilize the learnable queries to extract the 3D spatial information from text and multi-view images input, compared to models with the same number of parameters. It is reasonable as the MiniCPMV series employs cross-attention to map the image feature into the space of LLM rather than MLP, reducing the number of image tokens and resulting in the loss of comprehensive information. 2) Similarly, integration with Qwen2VL~\cite{wang2024qwen2vl} series demonstrates performance constraints. We argue that this limitation stems from the M-RoPE design of Qwen2VL, which causes the learnable queries to interact with fixed 2D position image features in LLM, leading to contradictions in subsequent 3D spatial understanding. 3) Implementations based on the InternVL2 series~\cite{chen2024internvl} exhibit progressive performance improvements as the model parameters increase, showcasing the trend of scaling law. In addition, InternVL2 exceeds InternVL1.5 in the same number of parameters, indicating that stronger initialization can improve model performance.

\textbf{Analysis on various 3D detectors as the spatial processor.} We then explore the effect of initializing the spatial processor from various 3D detectors. Specifically, we select two widely used 3D detector methods, i.e., BEV-based methods~\cite{li2022bevformer} and DETR-based methods~\cite{liu2022petr,xiong2023cape}. As listed in Tab.~\ref{tab:3d_component}. We first use the BEVFormer\footnote{Note that we only take current frame information instead of the default multi-frame with history information.}~\cite{li2022bevformer}, a BEV-based 3D detector, achieving suboptimal performance, where the mAP is 30.87\% and F1 score is 17.85\%. When we utilize PETR~\cite{liu2022petr}, a DETR-based 3D detector, we observe a notable improvement, where mAP is from 30.87\% to 34.26\%, and F1 score is from 17.85\% to 20.48\%. However, we observe a significant performance drop when we try to employ a more advanced DETR-based 3D detector, namely CAPE~\cite{xiong2023cape}. We argue the following reasons: 1) CAPE changes the global 3D position embedding (3D PE) to local, conflicting with the comprehensive understanding ability of LVLM. 2) CAPE introduces the denoise technique, breaking the accurate textual information from LLM.

\textbf{The influence of learnable query numbers.} An essential component of our framework design is addressing the 3D VG challenge. To this end, we introduce a set of learnable queries to bridge the pre-trained 3D detector and LVLM. Hence, we further investigate the influence of learnable query numbers. As shown in Tab.~\ref{tab:query_num_ablation}, increasing the number of queries from 10 to 900 leads to a 14.53\% improvement in Pr, while the mAP and F1 score decrease by 27.14\% and 15.96\%, respectively. 
This trend can be explained by the nature of the evaluation metrics. Since Pr is computed via one-to-one matching with GT, a higher number of queries increases the probability of correct matches, thereby improving recall. However, this also results in a larger number of total predictions, diluting the proportion of correct matches and leading to lower mAP. Additionally, as the learnable queries must first be processed by the LLM, increasing their number significantly raises computational overhead. Based on this trade-off, we set the number of queries to 30 by default.

\begin{table}[!t]
\centering
    \footnotesize
    \caption{The influence of 3D position embedding on 3D Visual Grounding tasks.}
    \label{tab:3d_pe_ablation}
    \setlength{\tabcolsep}{2.0mm} 
     \begin{tabular}{c c c c c c c c}
        \toprule
         3D PE& Pr@0.5   & Pr@1 & Pr@2 & Pr@4  & Pr  &mAP & F1\\
        \midrule
        w/o & 9.12 & 26.71 & 54.88 & 77.82 & 42.13 &\ 21.10 &12.87\\
        w   & \textbf{19.95}  & \textbf{40.78}  & \textbf{64.31}  & \textbf{81.61}  & \textbf{51.66} & 
        \textbf{34.26} &\textbf{20.48} \\
        \bottomrule
    \end{tabular}
\end{table}

\begin{table}[!t]
\centering
    \footnotesize
    \caption{The influence of query initialization on 3D Visual Grounding tasks.}
    \label{tab: query initialization}
    \setlength{\tabcolsep}{2.0mm} 
     \begin{tabular}{c c c c c c c c}
        \toprule
         Init & Pr@0.5   & Pr@1 & Pr@2 & Pr@4  & Pr  &mAP & F1\\
        \midrule
        Average & 18.72 & 38.55& 60.48& 77.44& 48.80&  25.50 &15.36\\
        Max   & 18.75 & 38.87 & 61.06& 78.77& 49.36 & 30.91& 18.59\\
        Random   & \textbf{19.95}  & \textbf{40.78}  & \textbf{64.31}  & \textbf{81.61}  & \textbf{51.66} & 
        \textbf{34.26} &\textbf{20.48} \\
        \bottomrule
    \end{tabular}
\end{table}

\begin{figure}[!t]
  \centering
  \includegraphics[width=\columnwidth]{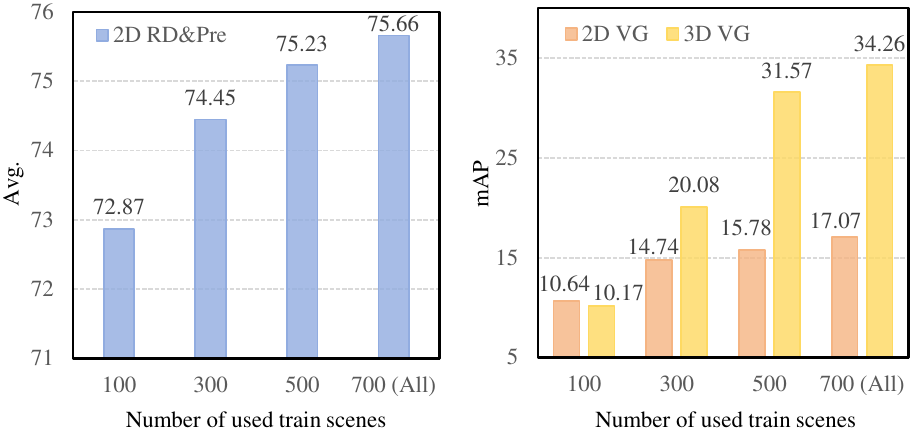}
  \caption{Comparison on all interactive tasks of using different data
size for training. Avg. denotes the average of BLEU, ROUGE, and CIDEr. RD: Region Description; Pre: Prediction; VG: Visual Grounding.}
  \label{fig:data_size_abalation}
  \vspace{-10pt}
\end{figure}

\textbf{The influence of 3D position embedding.} We subsequently examine the influence of 3D position embedding, providing 3D spatial priors and converting multi-view image features to 3D space features.
The incorporation of 3D PE yields substantial performance enhancements for the 3D VG task, as listed in
Table~\ref{tab:3d_pe_ablation}. Specifically, we observe significant improvements of 9.53\% and 13.16\% in Pr and mAP, respectively. These results underscore the importance of spatial context in establishing effective correspondences between visual elements and language descriptions within the 3D environment.

\begin{figure*} 
  \centering
 \includegraphics[width=\textwidth]{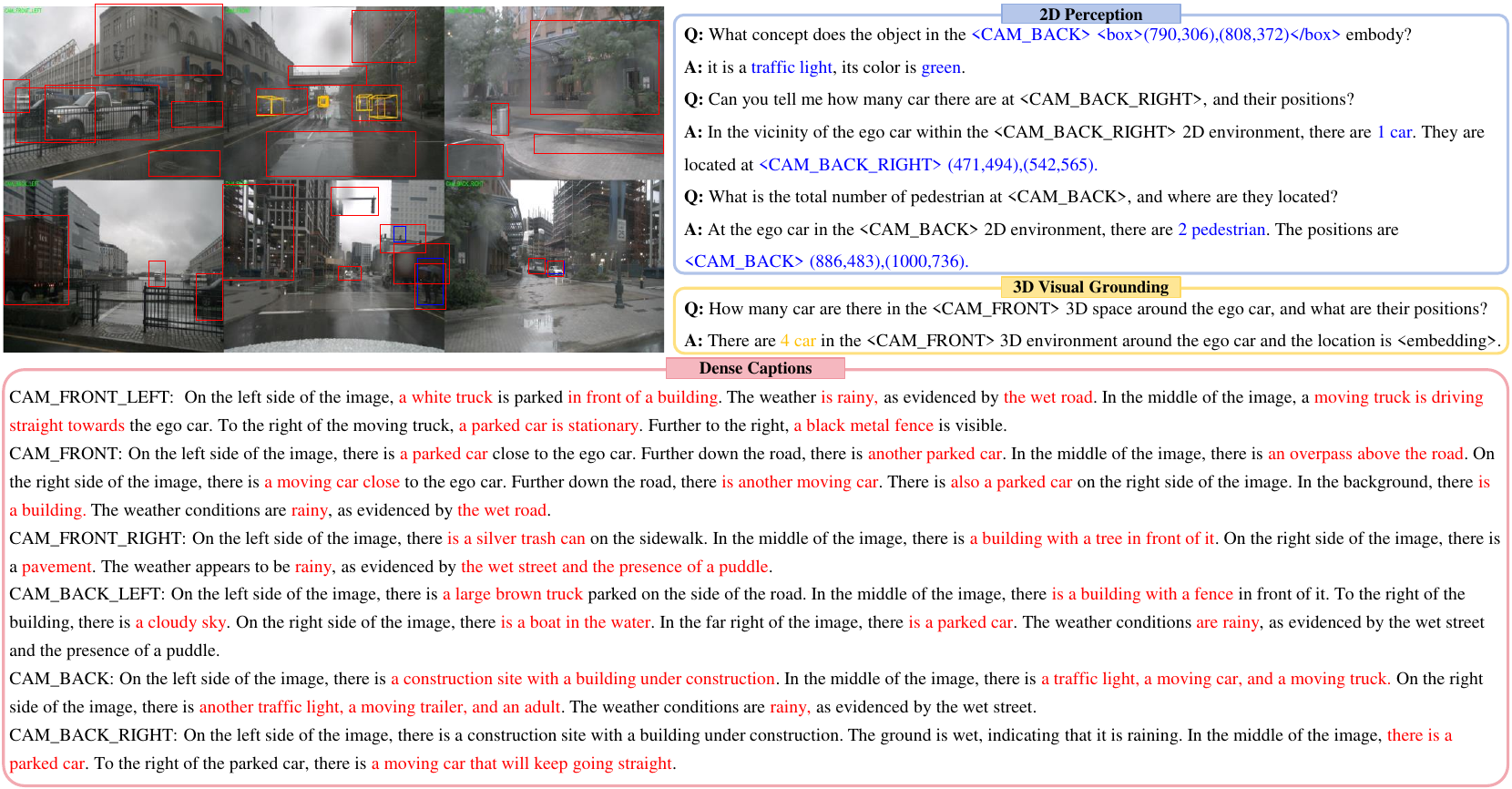}
  \caption{Qualitative results of Drivemonkey on the NuInteract test set. The \textcolor{red}{red}, \textcolor{blue}{blue}, and \textcolor{orange}{orange} refer to the captured object and corresponding description in the dense caption, the referred objects position and expression on 2D perception tasks, and predicted 3D bounding boxes on 3D VG tasks, respectively.}
  \label{fig: vis}
  \vspace{-10pt}
\end{figure*}

\textbf{The effect of query initialization mode.} To investigate the impact of query initialization strategies, we evaluate three distinct approaches: random initialization and initializing from image features via average pooling or max pooling. As shown in Tab.~\ref{tab: query initialization}, random initialization demonstrates superior performance, exceeding average pooling by 8.76\% on the mAP metric. This empirical finding suggests that direct utilization of image features does not necessarily contribute to more effective 3D priors. The observed performance differential indicates that utilizing an unbiased initialization state may facilitate more robust feature extraction in the 3D VG task rather than constraining it with potentially suboptimal image-derived initialization.

\textbf{The effect of data size.} Finally, we explore the relationship between training data volume and model performance, as shown in Fig.~\ref{fig:data_size_abalation}. Note that we use the average of BLEU, ROUGE, and CIDEr as the language metric. When the number of train scene data is relatively small, DriveMonkey exhibits performance degradation. However, as the training dataset expands, we observe consistent performance improvements across all interactive tasks. Specifically, the language metric of 2D region description \& prediction increases by 2.79\%, while the mAP of 2D VG task improves by 6.43\%. Most notably, the 3D VG task demonstrates remarkable enhancement with a 24.09\% increase in mAP, underscoring the exceptional scaling capabilities of our proposed framework.

\subsection{Qualitative results}

Fig.~\ref{fig: vis} presents qualitative results of Drivemonkey, showcasing its performance across multiple tasks, including driving scene description, 2D perception, and 3D VG tasks on the NuInteract test set. We observe that DriveMonkey can capture comprehensive elements in driving scenarios, such as buildings, traffic lights, and the footbridge, surpassing the annotation range of nuScene~\cite{caesar2020nuscenes}. Additionally, Drivemonkey accurately identifies the location and relevant attributes of the designated object, demonstrating surprising instruction-following capabilities and interactive responsiveness. These results highlight the benefits of our framework's seamless integration of the LVLM with a spatial processor. This design not only generates accurate textual responses but also ensures precise and high-quality performance in the 3D VG task.

\section{Conclusion}

In this paper, we present NuInteract, a large-scale and richly annotated multi-view image-language dataset tailored for comprehensive scene understanding in autonomous driving. It comprises over 1.5M image-language pairs across diverse tasks, including dense caption, perception, prediction, planning, and 3D visual grounding (VG), offering fine-grained contextual and spatial information beyond existing language-based driving datasets. We propose DriveMonkey, a flexible framework that seamlessly integrates the LVLM with a spatial processor via a set of learnable queries. The spatial processor, as a plug-and-play enhancement, enhances the LVLM's spatial understanding ability and boosts its performance in 3D perception. Extensive experiments demonstrate that DriveMonkey significantly outperforms existing LVLMs on our NuInteract benchmark, especially on the challenging 3D VG task, and shows notable gains after our dense caption pre-training. Our framework provides a novel paradigm for leveraging LVLM to perform 3D VG tasks, which may offer insights and serve as a reference for future research in the community.

\noindent\textbf{Limitation.} This paper primarily focuses on current moment comprehensive information for a given scene, without incorporating temporal information. However, temporal information is also crucial for autonomous driving systems to understand the scenario and make the correct decision. Additionally, our planning tasks only concentrate on high-level commands rather than specific trajectories, without comparison with advanced end-to-end autonomous driving systems. We would like to fill these gaps in future work.

\bibliographystyle{IEEEtran}
\bibliography{main}

\end{document}